\pdfoutput=1
\documentclass[11pt]{article}
\usepackage{amsfonts}
\usepackage{dashrule}
\usepackage{acl}
\usepackage{booktabs, siunitx}
\usepackage{microtype}
\usepackage{hyperref}
\usepackage{fancyhdr}
\usepackage{arydshln}
\usepackage{url}
\usepackage{booktabs}
\usepackage{makecell}
\usepackage{appendix}
\usepackage{mdframed}
\usepackage{tabularx}
\usepackage{pgfplots}
\usepackage{caption}
\pgfplotsset{compat=1.18}
\usepackage[breakable]{tcolorbox}
\usepackage{algorithm}
\usepackage{algorithmicx}
\usepackage{algpseudocode}
\usepackage{amsmath}
\usepackage{bbm}
\usepackage{booktabs}
\usepackage{multirow} 
\usepackage{colortbl}
\usepackage{times}
\usepackage{arydshln}
\usepackage{booktabs}
\usepackage{xcolor}
\usepackage{colortbl}
\usepackage{placeins}
\usepackage{multicol}
\usepackage{setspace}
\usepackage{afterpage}
\usepackage{float}
\usepackage{latexsym}
\usepackage{graphicx}
\usepackage{setspace} 
\usepackage{etoolbox}
\usepackage{enumitem}

\definecolor{darkblue}{rgb}{0, 0, 0.5}
\hypersetup{colorlinks=true, citecolor=darkblue, linkcolor=darkblue, urlcolor=darkblue}

\title{Efficient Dynamic Clustering-Based Document Compression\\ for Retrieval-Augmented-Generation}

\author{
\textbf{Weitao Li$^{1,2}$\thanks{Email: liwt23@mails.tsinghua.edu.cn}}, \textbf{Xiangyu Zhang$^{1}$}, \textbf{Kaiming Liu$^{1,2}$}, \textbf{Xuanyu Lei$^{1,2}$}, \textbf{Weizhi Ma$^{2,\dagger}$}, \textbf{Yang Liu$^{1,2,\dagger}$}\\
$^1$ Dept. of Comp. Sci. \& Tech., Institute for AI, Tsinghua University, Beijing, China \\
$^2$ Institute for AI Industry Research (AIR), Tsinghua University, Beijing, China \\
}

\begin{document}
\thispagestyle{fancy}
\fancyhf{}
\rhead{Accepted to EMNLP 2025 Findings}

\maketitle
{\let\thefootnote\relax\footnotetext{$\dagger$ corresponding authors.}}

\begin{abstract}
Retrieval-Augmented Generation (RAG) has emerged as a widely adopted approach for knowledge injection during large language model (LLM) inference in recent years. However, due to their limited ability to exploit fine-grained inter-document relationships, current RAG implementations face challenges in effectively addressing the retrieved noise and redundancy content, which may cause error in the generation results. To address these limitations, we propose an \textbf{E}fficient \textbf{D}ynamic \textbf{C}lustering-based document \textbf{C}ompression framework (\textbf{EDC\textsuperscript{2}-RAG}) that utilizes latent inter-document relationships while simultaneously removing irrelevant information and redundant content. We validate our approach, built upon GPT-3.5-Turbo and GPT-4o-mini, on widely used knowledge-QA and Hallucination-Detection datasets. Experimental results show that our method achieves consistent performance improvements across various scenarios and experimental settings, demonstrating strong robustness and applicability.
Our code and datasets are available at \url{https://github.com/Tsinghua-dhy/EDC-2-RAG}. 
\end{abstract}

\section{Introduction}
In recent years, large language models (LLMs) have advanced rapidly, excelling in natural language processing (NLP) tasks such as question answering, code generation, and even medical diagnosis~\cite{yasunaga-etal-2021-qa,he2025g,yue2023disc,singhal2023towards,li2024agent}. Despite their success, LLMs face two key challenges: expensive knowledge updates due to the large number of learnable parameters, and hallucinations that lead to misleading content~\cite{honovich2023unnatural,hu2023large,lin2024towards,xu2024hallucination}. These issues impact the availability, reliability and consistency of LLMs~\cite{zhou2024larger}. Retrieval-augmented generation (RAG)~\cite{lewis2020retrieval,borgeaud2022improving,izacard2022few} addresses these problems by integrating retrieval with generation, allowing LLMs to access external knowledge without parameter updates, reducing hallucinations, and improving reliability.

However, the implementation of RAG methods in real-world settings presents significant challenges. From a structural perspective, the effectiveness of RAG frameworks derives from the information augmentation of integrated databases\citep{lewis2020retrieval}. In practical applications, the databases are often of limited quality due to the scarcity of high-quality data and the high cost of data cleaning. Therefore, the candidate documents faced by retrievers tend to exhibit the following frequently-encountered quality flaws:

\begin{itemize}[leftmargin=1.5em,itemsep=0pt,parsep=0.2em,topsep=0.1em,partopsep=0.0em]
    \item \textbf{Noise}: irrelevant content to the query, which may result in errors during generation.
    \item \textbf{Redundancy}: highly similar content between documents, which will consume more tokens and time in inference.
\end{itemize}

\begin{figure*}[t]
     \centering
    \includegraphics[width=0.9\linewidth]{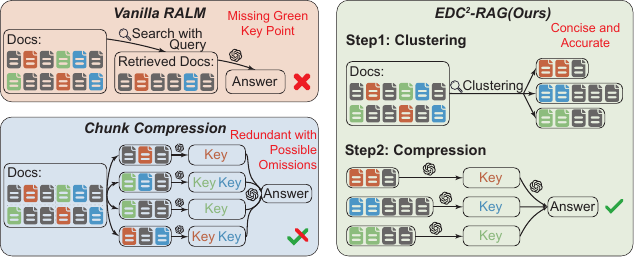}
     \caption{Comparison between our method and prior approaches. Unlike Vanilla RAG, which misses key information, and Chunk Compression, which is redundant and incomplete, our method clusters and compresses documents to extract concise and accurate answers.}
     \label{fig:1}
\end{figure*}

These issues can significantly reduce the effectiveness of retrieval and compromise the quality of the final generated output. Faced with these practical challenges, it is increasingly significant to build a reliable RAG system. However, current RAG frameworks predominantly rely on query-document similarity for retrieval, without explicitly addressing prevalent issues such as noise and redundancy in real-world document corpora. To solve the problems, we propose an efficient dynamic clustering-based compression method for a reliable document retrieval.

Specifically, we first encode the documents to get a denser content representation, then perform clustering to aggregate semantically similar documents, mitigating content repetition. Subsequently, we use prompt-based techniques to guide the LLMs in query-specific compression to improve information density and eliminate noise. Finally, we concatenate the compressed content into the prompts for response generation.
In summary, our method leverages the latent relationships between documents to reduce noise and redundant content.


To validate the effectiveness of our approach, we selected two types of widely used datasets: KQA tasks and hallucination detection tasks. Systematic experiments conducted on GPT-3.5-Turbo demonstrate that our method achieves significant performance improvements across different settings. Meanwhile, our method also exhibits strong robustness and generalization potential to other scenarios. These findings indicate that by deeply exploring and utilizing fine-grained relationships among documents, RAG methods can reach new performance heights, providing a novel direction for addressing the hallucination problem and knowledge update challenges in LLMs.

The main contributions of our work are:
\begin{itemize}[leftmargin=1.5em,itemsep=0pt,parsep=0.2em,topsep=0.1em,partopsep=0.0em]
\item To the best of our knowledge, we are the first to apply similarity-based semantic clustering in the post-retrieval stage to address practical challenges in in-the-wild RAG systems.
\item Our method effectively improves the performance and robustness of the RAG systems and also enhances their long context capability.
\item As a post-retrieval method, our approach is plug-and-play, requiring no additional training, and can be integrated into various pipelines.
\end{itemize}

\section{Related Works}

\paragraph{Reranking and Compression.}
Post-retrieval methods for frozen large language models (LLMs) can be categorized into reranking and compression approaches~\cite{gao2023retrieval}.
Reranking refines the order of retrieved documents to improve LLMs-generation performance.
Re3val~\cite{song-etal-2024-re3val} uses reinforcement learning (RL) and targeted queries, while REAR~\cite{wang-etal-2024-rear} utilizes LLaMA~2~\cite{touvron2023llama} for reranking, enhancing response quality.
Compression methods condense retrieved content, primarily through fine-tuned models\citep{xu2023recomp,liu-etal-2023-tcra,yu-etal-2024-chain} or LLMs native capabilities.
For instance, SURE~\cite{kim2023sure} generates and selects the best answer by summarizing multiple responses.
However, existing methods \textbf{rarely address document noise and redundancy issues}, whereas our approach tackles them with dynamic clustering and prompt-guided compression.

\paragraph{Retrieval Semantic Relation Modeling.}

\newmdenv[
  backgroundcolor=gray!5,
  linecolor=black!60,
  linewidth=0.5pt,
  topline=true,
  bottomline=true,
  leftline=false,
  rightline=false,
  skipabove=\baselineskip,
  skipbelow=\baselineskip,
  innerleftmargin=10pt,
  innerrightmargin=10pt,
  innertopmargin=10pt,
  innerbottommargin=10pt
]{grayboxalgo}

\makeatletter
\newcommand\mediumsmall{%
  \@setfontsize\mediumsmall{9.5pt}{11pt} 
}
\makeatother

\begin{figure*}[ht]
\centering
\begin{minipage}{0.95\textwidth}
\mediumsmall
\setstretch{0.97} 

\begin{grayboxalgo}
\begin{minipage}[t]{0.48\textwidth}
\raggedright
\textbf{Phase 1: Initialization}
\vspace{0.5em}
\begin{algorithmic}[1]
\State \textbf{Input:} Document set $V = \{d_1, d_2, \dots, d_n\}$, query $q$, similarity function $\text{sim}(\cdot,\cdot)$, embedding model $f(\cdot)$, initial cluster size $\tau$, threshold $\Lambda$
\State \textbf{Output:} Clusters $\{C_1, C_2, \dots, C_k\}$
\State Compute query embedding: $\mathbf{v}_q \gets f(q)$
\ForAll{$d_j \in V$}
  \State Compute embedding: $\mathbf{v}_j \gets f(d_j)$
\EndFor
\State Select initial cluster root: $C.R_1 \gets \arg\max_{d \in V} \text{sim}(\mathbf{v}_q, \mathbf{v}_j)$
\ForAll{$d_j \in V$}
  \State Compute similarity: $s_j \gets \text{sim}(\mathbf{v}_{C.R_1}, \mathbf{v}_j)$
\EndFor
\State Form $C_1$ with top-$\tau$ documents from $V$ sorted by $s_j$
\State Remove $C_1$ members from $V$
\end{algorithmic}
\end{minipage}
\hfill
\begin{minipage}[t]{0.48\textwidth}
\raggedright
\textbf{Phase 2: Iterative Subgraph Formation}
\vspace{0.5em}
\begin{algorithmic}[1]
\State $k \gets 2$
\While{$V \neq \emptyset$}
  \State Select new root: $C.R_k \gets \arg\max_{d \in V} \text{sim}(\mathbf{v}_q, \mathbf{v}_j)$
  \ForAll{$d_j \in V$}
    \State Compute similarity: $s_j \gets \text{sim}(\mathbf{v}_{C.R_k}, \mathbf{v}_j)$
  \EndFor
  \State Determine cluster size: $size \gets \min(2 \times |C_{k-1}|, \Lambda)$
  \State Form $C_k$ with top-$size$ documents from $V$ sorted by $s_j$
  \State Remove $C_k$ members from $V$
  \State $k \gets k + 1$
\EndWhile
\end{algorithmic}
\end{minipage}
\end{grayboxalgo}

\vspace{-0.5em}
\captionof{algorithm}{Efficient Dynamic Graph-based Document Clustering}
\label{alg:document_clustering}

\end{minipage}
\end{figure*}

Beyond post-retrieval methods, some studies focus on refining relationships between documents, chunks or entities.
Recent approaches frame RAG as a multi-agent collaboration, where each agent processes a subset of retrieved content. Long Agent~\cite{zhao-etal-2024-longagent} supports large contexts through chunk-level conflict resolution, while MADAM-RAG~\cite{wang2025retrieval} uses agents to address conflicting responses. Multi-agent RAG is also applied to data integration~\cite{salve2024collaborative}, but these methods increase inference costs and latency, limiting real-world applicability.
Knowledge Graphs (KGs) structure document information by providing contextual relationships~\cite{ji2021survey}. KAPING builds a KG for retrieval~\cite{baek-etal-2023-knowledge-augmented}, while G-Retriever queries subgraphs~\cite{he2025g}. Despite their effectiveness in entity-rich tasks, KG-based methods \textbf{face scalability and adaptability challenges} and often \textbf{require substantial resources on the corpus processing side}~\cite{peng2023knowledge, li2024simple}, and so does RAPTOR~\cite{sarthi2024raptor}.
 Our method dynamically constructs semantic relationships post-retrieval, avoiding multi-agent systems and pre-built graphs, thereby improving retrieval quality by reducing redundancy and noise.

\section{Problem Definition}

Consider a set of retrieved documents \( V = \{d_1, d_2, \ldots, d_n\} \), where each document \( d_i \) is associated with a query \( q \). These documents are retrieved based on their relevance to \( q \), but their exact utility in answering \( q \) is initially unknown. Furthermore, there may exist potential overlaps and redundancies among the documents in \( V \), as some documents may share similar or identical information, while others may provide complementary or conflicting details. 

Let \( E = \{e_{ij}\} \) represent the relationships between pairs of documents \( d_i \) and \( d_j \), where \( i, j \in \{1, 2, \ldots, n\} \). These relationships can be categorized as:
\begin{itemize}[leftmargin=1.5em,itemsep=0pt,parsep=0.2em,topsep=0.1em,partopsep=0.0em]
    \item \textbf{Overlapping}: \( e_{ij} = \text{Overlap} \), indicating that \( d_i \) and \( d_j \) share redundant or highly similar content.
    \item \textbf{Complementary}: \( e_{ij} = \text{Complementary} \), indicating that \( d_i \) and \( d_j \) provide distinct but relevant information to \( q \).
\end{itemize}

Additionally, let \( U = \{u_1, u_2, \ldots, u_n\} \) denote the utility scores of the documents, where \( u_i \) represents the degree to which \( d_i \) contributes to answering \( q \). These scores are initially unknown and must be inferred based on the relationships \( E \) and the content of the documents. 

The goal is to effectively utilize the retrieved documents \( V \), their relationships \( E \), and their inferred utilities \( U \) to construct a comprehensive and accurate response to the query \( q \). This involves addressing the challenges of redundancy, inconsistency, and varying utility among the documents, while ensuring that the final output maximizes relevance and minimizes noise.

\section{Method}  
\subsection{Overview}

The core of our approach involves clustering documents using embedding models guided by predefined rules, followed by applying compression techniques to eliminate noise. These refined documents are then integrated into the prompt, enabling the LLM to more effectively utilize the information and enhance its performance. Our methodology is presented in accordance with the processing workflow, and Figure~\ref{fig:1} provides a comparative visualization of our method against current RAG frameworks.

\subsection{Efficient Dynamic Clustering of Documents}  

In RAG frameworks, retrieved documents often contain redundancy and noise, which can negatively impact the reasoning quality of LLMs. Traditional post-retrieval methods primarily rely on reranking or compression strategies to refine retrieved results, but they often fail to fully utilize the fine-grained relationships between documents.

To address this, we propose an efficient dynamic clustering-based approach to structure the retrieved documents before further processing. By organizing documents into clusters based on similarity, we aim to reduce redundancy and group related content together, creating a more coherent input for downstream tasks. Specifically, we prioritize documents with high similarity to the query, as these are most likely to contribute valuable information. Additionally, we adopt a dynamically expanding clustering strategy, where the cluster size increases iteratively, ensuring efficient grouping while keeping computational costs manageable. In our experiments, \textbf{we set $\tau=3$ and $\Lambda=20$}.

\subsection{Query-Aware Compression}

After constructing the subgraphs $ C_1, C_2, \dots, C_k $, it is essential to further refine the retrieved content by eliminating redundancy and distilling key information. While clustering helps organize documents based on similarity, it does not inherently resolve the issue of overlapping or extraneous details.

To address this, we introduce a compression step that leverages a large language model (LLM) to generate concise yet informative summaries. Specifically, we concatenate each $C_i$ ($i \in [1, k]$) with the query $q$ and prompt the LLM to produce a query-aware summary, ensuring that only the most relevant and essential content is preserved. The goal of this step is to maximize the information density of retrieved documents while removing redundant or marginally relevant details, preparing a high-quality input for final generation. 

Importantly, this summarization process is highly efficient as \textbf{all summaries can be generated in parallel}, allowing the system to scale effectively with the number of clusters while maintaining low latency. An example prompt is as follows:

\begin{tcolorbox}[colback=gray!5, colframe=black!30, boxrule=0.5pt, title=\textbf{Compression Prompt}, breakable, fonttitle=\bfseries, fontupper=\small]
\noindent\makebox[\linewidth]{\hdashrule{\linewidth}{0.5pt}{1mm}}

\textbf{Few-shots:}\\
\texttt{\{example 1\}}\\[0.3em]
\texttt{\{example 2\}}\\[0.5em]
\texttt{\{...\}}\\[0.5em]
\textbf{Instruction}:\\
Given a \textcolor[HTML]{0000ff}{question} and a set of \textcolor[HTML]{0abf53}{reference documents}, extract only the \textbf{verifiable, relevant} information that directly supports the question.\\
Avoid inferences or conclusions.\\
If nothing is relevant, output: \texttt{"No content to extract"}.\\[0.5em]

\textbf{Question}:\\
\texttt{\{query\}}\\[0.5em]
\textbf{Documents}:\\
\texttt{\{docs\}}\\[0.5em]
\textbf{Extracted Summary}:\\
\texttt{\{to be filled\}}

\noindent\makebox[\linewidth]{\hdashrule{\linewidth}{0.5pt}{1mm}}
\end{tcolorbox}

\subsection{Generation}
After clustering and compression refine the documents, the system generates a contextually relevant response. Our query-aware integration ensures the output is based on coherent, information-rich content tailored to the query. To accommodate diverse dataset characteristics, our method flexibly adapts the generation process. In scenarios where compression may risk omitting critical details due to LLM limitations (such as in KQA tasks), we strategically integrate response generation with the compression phase, allowing the system to dynamically refine answers. This approach enhances the retention of essential information and improves response accuracy, particularly in complex question-answering tasks. If compression yields poor summaries, the system falls back to original documents, ensuring robustness.

Unlike traditional RAG methods, which often rely on loosely structured retrieved documents, our approach enhances the informativeness of retrieved content by distilling critical insights in a query-driven manner. This structured input enables the LLM to reason more effectively, reducing hallucinations and improving response precision. Moreover, our method efficiently balances computational costs and performance by limiting the number of API calls required for summarization, ensuring practical deployment feasibility.

By optimizing the input for the final response generation step, our method improves both the precision and efficiency of the system, leading to more reliable and contextually relevant outputs while reducing computational overhead.

\section{Experimental Settings}

\subsection{Overview}

To validate the effectiveness of our method, we employe three types of datasets in the experiments: Knowledge-QA datasets, Hallucination-Detection datasets, and Redundancy dataset built by us. The retrieval settings and implementation details for these datasets vary slightly, which are presented in Appendix~\ref{sec:implementation details}.  

We utilize GPT-3.5-Turbo-1106 and GPT-4o-mini-2024-07-18 as the backbone LLMs. For simplicity, we refer to GPT-3.5-Turbo-1106 as "ChatGPT" and GPT-4o-mini-2024-07-18 as "GPT-4o-mini". The decoding temperature is fixed at 0 for reproducibility, with the exception of Long Agent and KQA sampling steps in our methods, where 0.7 is used to enhance output diversity.

\subsection{Datasets}  

\textbf{Knowledge-QA Datasets}: Knowledge Question Answering (KQA) datasets assess a LLM’s ability to reason over retrieved external knowledge sources from knowledge graphs or textual corpora. We use three common KQA datasets~\cite{yu-etal-2024-chain, lv2024coarse, song2025r1}: WebQ~\cite{berant2013semantic} (single-hop), and 2WikiMultiHopQA~\cite{ho-etal-2020-constructing} (hereafter referred to as 2Wiki) plus Musique~\cite{trivedi-etal-2022-musique} (both multi-hop). To analyze noise robustness, following prior work~\cite{lv2024coarse, yu-etal-2024-chain}, we employ DPR retrieval and its reader to identify noisy documents, constructing cases with varying noise proportions by filtering samples from these three datasets. Details are in the Appendix~\ref{sec:knowledge-qa-datasets-intro}.

\textbf{Redundancy dataset}: To evaluate the capability of our method in handling redundancy, we used DPR to retrieve Top-$20$ documents per question from the WebQ dataset. The redundancy rate $r$ is defined as:

\vspace{-12pt}

\vspace{-3pt}

\[
r = \frac{\text{number of rewritten documents}}{20}
\]

\vspace{-3pt}

Implementation details are provided the in Appendix~\ref{sec:knowledge-qa-datasets-intro}.


\textbf{Hallucination-Detection Datasets}: Hallucination Detection is an NLP task that verifies whether generated or stated content---like summaries or answers---is factual or nonfactual by checking against available information sources. We conducte experiments on three widely used fact-checking tasks~\cite{li-etal-2024-citation,lv2024coarse}: the FELM World Knowledge Subset~\cite{chen2023felm}, the WikiBio GPT-3 Dataset~\cite{manakul-etal-2023-selfcheckgpt}, and the HaluEval Dataset~\cite{li-etal-2023-halueval}. Details are in the Appendix~\ref{sec:Hallucination-Detection-datasets-intro}.
\subsection{Baselines and Evaluation Metrics}
We compare with several baselines: 1) \textbf{Vanilla RALM}~\cite{borgeaud2022improving}, the standard RAG process; 2) \textbf{Chunk Compression}~\cite{jiang-etal-2024-longllmlingua}, which compresses documents using an LLM; 3) \textbf{Long Agent}~\cite{zhao-etal-2024-longagent}, which divides long documents among collaborating agents with a leader agent aggregating outputs; 4) \textbf{CEG}~\cite{li-etal-2024-citation}, a strong post-hoc RAG baseline for hallucination detection; 5) \textbf{Raptor}, which leverages recursive abstractive processing for tree-organized retrieval; and 6) task-specific methods including \textbf{HalluDetector}~\cite{wang-etal-2023-hallucination}, \textbf{Focus}~\cite{zhang-etal-2023-enhancing-uncertainty}, \textbf{SelfCheckGPT} w/NLI~\cite{manakul-etal-2023-selfcheckgpt}, CoT-augmented prompting~\cite{kojima2022large},  and prompts augmented with hyperlinks to reference documents and with human-annotated reference documents~\cite{chen2023felm}. Full details are in Appendix~\ref{sec:baselines}.

We use F1 score as the evaluation metric for the Knowledge-QA task, Balanced\_Acc for the FELM and WikiBio GPT-3 datasets, and Acc for the HaluEval dataset.

\section{Experimental Results}
\subsection{Main Results on Knowledge-QA Datasets}
\begin{table*}[h]
\centering
\small
\setlength{\tabcolsep}{6pt}
\renewcommand{\arraystretch}{1.2}
\begin{tabular}{p{2.0cm} p{3.4cm} c c c c c c c c}
\toprule
\textbf{Dataset} & \textbf{Method} 
& \multicolumn{8}{c}{\textbf{Top-$k$}} \\
\cmidrule(lr){3-10}
& 
& 5 & 10 & 20 & 30 & 50 & 70 & 100 & Avg \\
\midrule
\multicolumn{10}{c}{\textcolor{gray}{\textbf{gpt-3.5-turbo-1106}}} \\
\multirow{4}{*}{Musique}
& Vanilla RALM      & 71.05 & 71.73 & 74.75 & 76.93 & 75.16 & 80.25 & 77.04 & 75.27 \\
& Chunk Compression & 74.45 & 81.01 & 74.15 & 76.49 & 69.57 & 74.53 & 67.17 & 73.91 \\
& Long Agent        & \textbf{83.07} & \textbf{85.83} & \underline{82.04} & \textbf{84.84} & \underline{81.87} & \underline{80.65} & \underline{83.67} & \underline{83.14} \\
& \textbf{Ours}     & \underline{81.66} & \underline{83.31} & \textbf{82.55} & \underline{80.17} & \textbf{86.60} & \textbf{86.10} & \textbf{84.68} & \textbf{83.58} \\[2pt]
\hdashline
\multirow{4}{*}{WebQ}
& Vanilla RALM      & 88.84 & 90.14 & 90.07 & 90.30 & 91.13 & 90.74 & \underline{91.38} & \underline{90.89} \\
& Chunk Compression & \underline{90.52} & \textbf{91.15} & \underline{90.77} & \underline{91.18} & \underline{91.24} & \underline{90.98} & 90.38 & 90.26 \\
& Long Agent        & 89.79 & 91.03 & 90.49 & 90.25 & 89.01 & 90.21 & \underline{91.03} & 90.26 \\
& \textbf{Ours}     & \textbf{92.01} & 90.98 & \textbf{90.79} & \textbf{91.74} & \textbf{92.97} & \textbf{91.51} & \textbf{92.45} & \textbf{91.78} \\[2pt]
\hdashline
\multirow{4}{*}{2Wiki}
& Vanilla RALM      & \underline{69.90} & \textbf{74.68} & \textbf{77.51} & 71.36 & \underline{78.25} & 76.88 & 79.17 & 75.39 \\
& Chunk Compression & 67.38 & 67.14 & 72.41 & 68.98 & 72.08 & 72.99 & 72.66 & 70.52 \\
& Long Agent        & 69.30 & \underline{75.39} & 76.06 & \underline{78.36} & 77.16 & \textbf{83.22} & \textbf{83.45} & \underline{77.56} \\
& \textbf{Ours}     & \textbf{73.09} & \textbf{74.68} & \underline{76.20} & \textbf{78.64} & \textbf{80.90} & \underline{80.45} & \underline{82.06} & \textbf{78.00} \\
\midrule
\multicolumn{10}{c}{\textcolor{gray}{\textbf{gpt-4o-mini-2024-07-18}}} \\
\multirow{4}{*}{Musique}
& Vanilla RALM      & 74.43 & \underline{78.85} & 77.78 & 74.95 & 78.55 & 76.24 & 78.20 & 77.00 \\
& Chunk Compression & \underline{77.12} & 73.59 & 75.67 & \textbf{76.02} & 75.17 & 75.35 & \underline{79.42} & 76.05 \\
&RAPTOR &75.14&69.40 &72.07 &73.49 &\underline{78.65}&70.61&74.89&73.46\\
& Long Agent        & 73.29 & 75.25 & \underline{80.43} & 72.52 & \textbf{80.03} & \textbf{80.85} & 77.38 & \underline{77.11} \\
& \textbf{Ours}     & \textbf{78.33} & \textbf{79.80} & \textbf{81.71} & \underline{73.13} & 78.21 & \underline{77.95} & \textbf{80.07} & \textbf{78.46} \\[2pt]
\hdashline
\multirow{4}{*}{WebQ}
& Vanilla RALM      & 85.92 & 89.14 & 88.05 & 85.10 & 89.32 & \textbf{91.92} & 87.42 & 88.12 \\
& Chunk Compression & 85.64 & 84.99 & 85.07 & 83.98 & 88.66 & 90.79 & 90.94 & 87.15 \\
& Long Agent        & \underline{89.35} & \underline{89.16} & \underline{90.77} & \textbf{91.08} & \underline{91.82} & 90.91 & \underline{91.52} & \underline{90.66} \\
& \textbf{Ours}     & \textbf{90.01} & \textbf{90.77} & \textbf{91.89} & \underline{90.30} & \textbf{91.51} & \underline{91.25} & \textbf{92.02} & \textbf{91.11} \\[2pt]
\hdashline
\multirow{4}{*}{2Wiki}
& Vanilla RALM      & 64.81 & \textbf{73.38} & \textbf{73.84} & \underline{77.08} & \underline{78.04} & \textbf{78.01} & 77.89 & \underline{74.72} \\
& Chunk Compression & 62.38 & 65.76 & 69.24 & 67.62 & 72.45 & 73.26 & 74.06 & 69.25 \\
& Long Agent        & \underline{66.00} & \underline{70.04} & 71.33 & \textbf{77.68} & \textbf{79.98} & 77.13 & \textbf{83.45} & \textbf{75.09} \\
& \textbf{Ours}     & \textbf{68.67} & 69.79 & \underline{72.86} & 73.73 & 75.82 & \underline{77.43} & \underline{79.28} & 73.94 \\
\bottomrule
\end{tabular}
\caption{Performance comparison of different methods on MusiQue, WebQ, and 2Wiki Datasets Using GPT-3.5-turbo-1106 and GPT-4o-mini-2024-07-18 across various Top-$k$ values.}
\label{tab:top_k}
\end{table*}
\subsubsection{Results on Varying Top-\texorpdfstring{$k$}{k}}

Experimental results in Table~\ref{tab:top_k} demonstrate the effectiveness and robustness of our method across multiple datasets and LLM backends. 

On Musique, our approach achieves the highest average F1-scores with both ChatGPT and GPT-4o-mini, consistently outperforming all baselines. Notably, while Long Agent performs well with ChatGPT, its performance drops significantly with GPT-4o-mini, indicating possible overfitting or reduced adaptability. In contrast, our method maintains strong performance across both models.

On WebQ, our method also achieves the best average performance with ChatGPT and GPT-4o-mini, showing improvements over Vanilla RALM and other compression-based methods. The results highlight the generalizability of our approach to both simple and diverse question types.

For 2Wiki, a dataset requiring deeper reasoning, our method achieves the highest average with ChatGPT again, and shows competitive performance with GPT-4o-mini. Moreover, our approach exhibits more stable behavior across top-$k$ values, unlike some baselines that fluctuate significantly---especially Chunk Compression, whose performance is inconsistent across different $k$.

Overall, these results confirm that our clustering-based compression method is not only effective in preserving essential information and reducing redundancy, but also exhibits strong model-agnostic adaptability and stability across retrieval depths, making it a reliable choice for RAG pipelines.

\subsubsection{Results on Noise Resistence}

Tables~\ref{tab:noise_topk_100} and~\ref{tab:noise_topk_20} summarize performance under varying noise levels with Top-$k$ set to 100 and 20, respectively. Our method consistently yields the highest average F1 scores across all datasets and both model backends (ChatGPT and GPT-4o-mini). As noise increases, the performance gap over baselines widens, highlighting the robustness of our approach in noisy retrieval settings.

For instance, on MusiQue with ChatGPT at Top-$k$=100, our method exceeds the best baseline by over 3.4 F1 points on average and ranks first across all noise levels. Even at 100\% noise---when all retrieved documents are distractors---it achieves 84.54 F1, far surpassing the next-best score of 80.47. This demonstrates our compression strategy’s ability to suppress irrelevant content and recover useful signals from fully corrupted inputs.

Results on 2Wiki reveal similar strengths. While other methods degrade sharply with noise, our approach sustains relatively high performance, maintaining a 5–7 point margin under heavy noise. This shows its robustness in multi-hop reasoning even with deeply buried evidence.

GPT-4o-mini results show greater overall stabil-

\begin{table*}[ht]
\centering
\small
\setlength{\tabcolsep}{6pt}
\renewcommand{\arraystretch}{1.2}
\begin{tabular}{p{2.0cm} p{3.5cm} c c c c c c c}
\toprule
\textbf{Dataset} & \textbf{Method} 
& \multicolumn{7}{c}{\textbf{Noise Rates (\%) at Top-$k$=100}} \\
\cmidrule(lr){3-9}
&  & 0 & 20 & 40 & 60 & 80 & 100 & Avg \\
\midrule
\multicolumn{9}{c}{\textcolor{gray}{\textbf{gpt-3.5-turbo-1106}}} \\
\multirow{4}{*}{MusiQue} 
& Vanilla RALM & 77.04 & \underline{82.48} & \underline{79.32} & 76.49 & 79.45 & 75.86 & 78.44 \\
& Chunk Compression & 67.17 & 77.83 & 75.62 & \underline{79.79} & \underline{77.20} & 75.81 & 75.57 \\
& Long Agent & \underline{80.54} & 79.52 & 79.29 & \textbf{84.08} & \underline{77.20} & \underline{80.47} & \underline{80.18} \\
& \textbf{Ours} & \textbf{84.68} & \textbf{85.06} & \textbf{85.43} & 81.84 & \textbf{80.32} & \textbf{84.54} & \textbf{83.65} \\[2pt]
\hdashline

\multirow{4}{*}{WebQ}
& Vanilla RALM & \underline{91.38} & 88.88 & 88.28 & 88.85 & 87.54 & 81.61 & 87.76 \\
& Chunk Compression & 90.38 & 88.07 & 88.73 & \underline{89.73} & 87.10 & 82.87 & 87.81 \\
& Long Agent & 91.03 & \underline{90.79} & \underline{90.07} & 88.39 & \underline{90.17} & \underline{88.56} & \underline{89.84} \\
& \textbf{Ours} & \textbf{92.45} & \textbf{92.04} & \textbf{92.40} & \textbf{90.67} & \textbf{91.08} & \textbf{90.20} & \textbf{91.47} \\[2pt]
\hdashline
\multirow{4}{*}{2Wiki}
& Vanilla RALM & 79.17 & 71.76 & 71.48 & 71.26 & 64.81 & 58.95 & 69.57 \\
& Chunk Compression & 72.66 & 65.74 & 66.76 & 69.96 & 66.20 & 59.03 & 66.73 \\
& Long Agent & \textbf{83.45} & \textbf{81.41} & \textbf{82.52} & \textbf{78.88} & 71.79 & 70.92 & \textbf{78.16} \\
& \textbf{Ours} & \underline{82.06} & \underline{77.78} & \underline{74.69} & \underline{78.14} & \textbf{76.71} & \textbf{75.65} & \underline{77.51} \\
\midrule
\multicolumn{9}{c}{\textcolor{gray}{\textbf{gpt-4o-mini-2024-07-18}}} \\
\multirow{4}{*}{MusiQue}
& Vanilla RALM & 78.20 & 76.55 & 72.70 & 67.36 & \underline{76.49} & 64.94 & 72.71 \\
& Chunk Compression & \underline{79.42} & \underline{76.90} & \underline{75.62} & \underline{71.98} & 70.85 & 69.66 & 74.07 \\
& Long Agent & 77.38 & 75.93 & 74.76 & 73.44 & \textbf{76.58} & \textbf{78.84} & \underline{76.16} \\
& \textbf{Ours} & \textbf{80.07} & \textbf{82.17} & \textbf{77.49} & \textbf{74.43} & 75.62 & \underline{78.70} & \textbf{78.08} \\[2pt]
\hdashline
\multirow{4}{*}{WebQ}
&Vanilla RALM & 87.42 & 87.08 & \underline{89.67} & 85.13 & \textbf{90.31} & 84.89 & 87.42 \\
&Chunk Compression & \underline{90.94} & 90.06 & 89.30 & \underline{89.64} & 88.68 & 84.41 & 88.84 \\
&Long Agent & 91.77 & \underline{90.37} & \textbf{90.70} & \textbf{90.42} & 87.84 & \underline{86.67} & \underline{89.63} \\
&\textbf{Ours} & \textbf{92.02} & \textbf{91.42} & 89.31 & 88.97 & \underline{89.82} & \textbf{86.83} & \textbf{89.73} \\[2pt]
\hdashline
\multirow{4}{*}{2Wiki}
& Vanilla RALM & 77.89 & \underline{77.83} & \underline{75.79} & \textbf{77.15} & \textbf{72.69} & \underline{66.67} & \textbf{74.67} \\
& Chunk Compression & 74.06 & 75.19 & 75.58 & 73.88 & \underline{70.65} & 63.54 & 72.15 \\
& Long Agent & \textbf{83.45} & \textbf{81.13} & \textbf{76.97} & 73.99 & 64.06 & 59.64 & 73.21 \\
& \textbf{Ours} & \underline{79.28} & 76.27 & 75.35 & 71.96 & 70.64 & \textbf{68.67} & \underline{73.70} \\
\bottomrule
\end{tabular}
\caption{Comparison of F1 scores under different noise levels at Top-$k$=100 on MusiQue, WebQ, and 2Wiki datasets for multiple retrieval methods.}
\label{tab:noise_topk_100}
\end{table*}

\noindent ity than ChatGPT, but our method remains consistently superior. On MusiQue, it achieves 79.11 average F1, compared to 76.55 by Long Agent, again outperforming strong long-context baselines.

Under the Top-$k$=20 setting, where retrieval is constrained and noise more impactful, our method remains highly resilient. On WebQ and MusiQue, it sustains strong performance even under 80–100\% noise, while baselines drop sharply—demonstrating that our compression mechanism works effectively not only for large retrieval sets but also in low-budget scenarios where every document matters.

\subsubsection{Results on Redundancy Resistence}
Table~\ref{tab:webq_redundancy} reports performance under varying redundancy rates. Our method achieves the highest average F1 on WebQ, outperforming RALM in high-redundancy settings with a peak gain of +6.18 at 95\% redundancy. This demonstrates its effectiveness in handling redundant information while preserving retrieval quality.

In summary, our method’s consistent advantage across noise levels, datasets, and LLM backends highlights the generalizability and robustness of the compression strategy. By filtering irrelevant content and distilling key evidence, it boosts downstream performance and offers a reliable solution for noisy retrieval in RAG pipelines.

\begin{table*}[h]
\centering
\renewcommand{\arraystretch}{1.2}
\setlength{\tabcolsep}{5pt}
\scalebox{0.85}{
\begin{tabular}{@{}l l *{7}{>{\centering\arraybackslash}p{1.4cm}}@{}}
    \toprule
    Dataset & Method & \multicolumn{7}{c}{Redundancy Rates (\%) at Top-$k$=20} \\
    \cmidrule(lr){3-9}
     & & 0 & 20 & 40 & 60 & 80 & 95 & Avg \\
    \midrule
    \multirow{4}{*}{WebQ}
     & Vanilla RALM      & 90.07 & 87.67 & 89.76 & 89.00 & 88.17 & 83.04 & 87.95 \\
     & Chunk Compression & \underline{90.77} & 89.74 & \underline{90.21} & \underline{90.96} & 90.90 & 87.01 & 89.93 \\
     & Long Agent        & 90.25 & \textbf{92.31} & 88.75 & 88.98 & \textbf{90.95} & \textbf{89.89} & \underline{90.19} \\
     & \textbf{Ours}              & \textbf{92.01} & \underline{91.33} & \textbf{90.96} & \textbf{91.07} & \underline{90.93} & \underline{89.22} & \textbf{90.92} \\
    \bottomrule
\end{tabular}
}
\caption{Performance on WebQ under different redundancy rates (Top-$k$=20). Values in parentheses indicate differences from Vanilla RALM. Green indicates improvement, red indicates decline.}
\label{tab:webq_redundancy}
\end{table*}

\begin{table*}[h]
\centering
\renewcommand{\arraystretch}{1.2}
\setlength{\tabcolsep}{5pt}
\scalebox{0.85}{
\begin{tabular}{@{}l l *{7}{>{\centering\arraybackslash}p{1.4cm}}@{}}
    \toprule
    Dataset & Method & \multicolumn{7}{c}{Noise Rates (\%) at Top-$k$=20
} \\
    \cmidrule(lr){3-9}
     & & 0 & 20 & 40 & 60 & 80 & 100 & Avg \\
    \midrule
    \multirow{3}{*}{WebQ}
     & Dynamic & 90.79 & 91.87 & 90.75 & 91.00 & 89.23 & 87.87 & \textbf{90.25} \\
     & Avg     & 88.94 & 89.07 & 89.92 & 86.80 & 86.53 & 86.96 & 88.04 \\
     & Random  & 90.40 & 86.84 & 85.81 & 86.81 & 87.78 & 88.19 & 87.64 \\
    \bottomrule
\end{tabular}
}
\caption{Ablation study on clustering strategies under varying noise rates on WebQ.}
\label{tab:clustering}
\end{table*}

\subsection{Main Results on Hallucination Detection}
Table~\ref{tab:performance} presents a performance comparison of our proposed method against baseline approaches across three Hallucination-Detection datasets: FELM, WikiBio, and HaluEval. Results are reported as Maximum and Average accuracy over Top-$k$ predictions ($k$ from 1 to 10), with balanced accuracy used for FELM and WikiBio, and standard accuracy for HaluEval. Improvements over the best baseline are highlighted in green.

In the FELM dataset, our method achieves the highest maximum accuracy, surpassing baselines like Vanilla, CoT, Link. Our method performs only slightly below Doc, which benefits from manually annotated golden documents. Its average accuracy reflects a modest improvement over the CEG baseline, demonstrating robustness across varying $k$ values. For WikiBio GPT-3, our method performs competitively, slightly improving average accuracy over CEG and outperforming HalluDetector, Focus, and SelfCheckGPT, indicating consistent detection in biographical data. In HaluEval, our method records the highest performance, with a notable improvement over CEG, showcasing its effectiveness in open-domain settings.

\begin{table}[!h]
    \small
    \centering
    \sisetup{table-format=2.2}
    \begin{tabular}{@{} p{1.5cm} p{2.4cm} l @{}}
    \toprule
    \multicolumn{1}{c}{\textbf{Dataset}} & 
    \multicolumn{1}{c}{\textbf{Methods}} & 
    \multicolumn{1}{c}{\makecell{\textbf{Accuracy} \\ \textbf{(Top-$k$, $k{=}1{\sim}10$)}}} \\
    \midrule

    \multirow{6}{*}{FELM} 
     & Vanilla & 58.18 \\
     & CoT     & 61.32 \\
     & Link    & 56.78 \\
     & Doc     & \textbf{65.18} \\
     & CEG     & 63.35 / 61.89 \\
     & \textbf{Ours} & \underline{64.03} / 62.26\textsuperscript{\color{green!60!black}{+0.37}} \\
    \midrule

    \multirow{5}{*}{WikiBio} 
     & HalluDetector & 74.82 \\
     & Focus         & 74.08 \\
     & SelfCheckGPT  & 70.55 \\
     & CEG           & \textbf{76.58} / 74.14 \\
     & \textbf{Ours} & \underline{75.89} / 74.29\textsuperscript{\color{green!60!black}{+0.15}} \\
    \midrule

    \multirow{2}{*}{HaluEval} 
     & CEG           & 78.10 / 76.93 \\
     & \textbf{Ours} & \textbf{78.85} / 77.87\textsuperscript{\color{green!60!black}{+0.94}} \\
    \bottomrule
    \end{tabular}
    
    \caption{Performance comparison on Hallucination-Detection datasets. Each entry shows Max / Avg accuracy over Top-$k$. Metric: Accuracy for HaluEval; Balanced Accuracy for WikiBio GPT-3 and FELM.}
    \label{tab:performance}
\end{table}

Overall, our method consistently outperforms or matches the best baselines across all datasets, with improvements in average accuracy. These results highlight its stability and generalizability, making it a promising approach for reducing hallucinations in applications like automated fact-checking.

\subsection{Effectiveness of Clustering Strategies}

To validate the effectiveness of our clustering method, we compare it with two alternative strategies---Average Clustering and Random Clustering---that match our dynamic clustering in both the number of clusters and the overall document compression ratio for a controlled comparison. Average Clustering groups documents by their similarity rank to the query and distributes them evenly across clusters, while Random Clustering assigns documents randomly from the top-$k$ pool, maintaining the same number and size of clusters as dynamic clustering.

Table~\ref{tab:clustering} compares these strategies on WebQ under different noise rates. Our method achieves highest average F1, outperforming baselines. Average Clustering and Random Clustering obtain lower F1, and degrade more under high noise. These results highlight the effectiveness of our entropy-guided dynamic clustering in document compression.

Further validation is provided by evaluating clustering consistency on the Musique dataset using GPT-4o-mini-2024-07-18 for document classification. We measure the intra-class clustering probability for documents labeled as ``useful'' or ``noise,'' 

\begin{table*}[ht]
\centering
\small
\renewcommand{\arraystretch}{1.2}
\setlength{\tabcolsep}{1.2pt}
\begin{tabular}{l c c c c c c c c c c c}
\toprule
\multicolumn{12}{c}{Top-$k=20$} \\
\midrule
$\tau$ & 1 & 2 & \textbf{3} & 4 & 5 & 6 & 7 & 8 & 9 & 10 & 20 \\
\midrule
API Calls & 5 & 4 & 3 & 3 & 3 & 2 & 2 & 2 & 2 & 2 & 1 \\
F1 (\%) & 
72.86\textsubscript{\textcolor{gray}{±1.62}} &
73.07\textsubscript{\textcolor{gray}{±2.40}} &
76.85\textsubscript{\textcolor{gray}{±1.98}} &
77.15\textsubscript{\textcolor{gray}{±2.89}} &
74.70\textsubscript{\textcolor{gray}{±0.09}} &
76.69\textsubscript{\textcolor{gray}{±3.74}} &
76.51\textsubscript{\textcolor{gray}{±2.11}} &
74.88\textsubscript{\textcolor{gray}{±1.82}} &
77.63\textsubscript{\textcolor{gray}{±0.82}} &
73.55\textsubscript{\textcolor{gray}{±3.42}} &
73.71\textsubscript{\textcolor{gray}{±1.93}} \\
\midrule
\multicolumn{12}{c}{Top-$k=100$} \\
\midrule
$\tau$ & 1 & 2 & \textbf{3} & 4 & 5 & 6 & 7 & 8 & 9 & 10 & 20 \\
\midrule
API Calls & 7 & 6 & 6 & 5 & 5 & 5 & 4 & 4 & 4 & 4 & 3 \\
F1 (\%) & 
78.54\textsubscript{\textcolor{gray}{±3.74}} &
78.33\textsubscript{\textcolor{gray}{±1.70}} &
80.73\textsubscript{\textcolor{gray}{±3.90}} &
80.21\textsubscript{\textcolor{gray}{±3.12}} &
76.66\textsubscript{\textcolor{gray}{±3.32}} &
76.86\textsubscript{\textcolor{gray}{±2.14}} &
76.80\textsubscript{\textcolor{gray}{±2.02}} &
77.41\textsubscript{\textcolor{gray}{±2.69}} &
77.85\textsubscript{\textcolor{gray}{±1.17}} &
77.78\textsubscript{\textcolor{gray}{±2.14}} &
77.97\textsubscript{\textcolor{gray}{±2.09}} \\
\bottomrule
\end{tabular}
\caption{Ablation study results on Musique dataset (GPT-4o-mini-2024-07-18) for varying $\tau$ at top-$k=20$ and top-$k=100$ (noise = 40\%).}
\label{tab:ablation-musique}
\end{table*}

\noindent defined as:
\[
\frac{\sum_{i,j \in \text{same-class}, i<j} \mathbbm{1}[\text{cluster}(i) = \text{cluster}(j)]}{\binom{N_{\text{same-class}}}{2}}
\]

Table~\ref{tab:musique-consistency} summarizes these metrics under varying top-$k$ and noise levels, with random baselines using the same number of clusters. Our method exhibits probabilities exceeding random baselines, demonstrating significant semantic consistency and robustness, particularly under high noise.

\begin{table}[h]
\centering
\small
\renewcommand{\arraystretch}{1.1}
\setlength{\tabcolsep}{2.5pt}
\begin{tabular}{@{}l c c c c@{}}
\toprule
\multicolumn{5}{c}{Noise Rates (\%) at Top-$k=20$} \\
\midrule
Metric & 20 & 40 & 60 & 80 \\
\midrule
Useful Prob. (\%) & 35.87 & 36.59 & 36.43 & 39.37 \\
Rand. Useful (\%) & 33.33 & 33.33 & 33.33 & 33.33 \\
Noise Prob. (\%) & 31.43 & 34.97 & 35.22 & 35.05 \\
Rand. Noise (\%) & 33.33 & 33.33 & 33.33 & 33.33 \\
\midrule
\multicolumn{5}{c}{Noise Rates (\%) at Top-$k=100$} \\
\midrule
Metric & 20 & 40 & 60 & 80 \\
\midrule
Useful Prob. (\%) & 19.09 & 20.49 & 18.80 & 19.11 \\
Rand. Useful (\%) & 14.56 & 14.62 & 14.29 & 14.29 \\
Noise Prob. (\%) & 20.31 & 20.19 & 17.35 & 17.03 \\
Rand. Noise (\%) & 14.56 & 14.62 & 14.29 & 14.29 \\
\bottomrule
\end{tabular}
\caption{Clustering consistency metrics on Musique dataset (GPT-4o-mini-2024-07-18 classification) under varying top-$k$ and noise levels, displayed for Top-$k=20$ and Top-$k=100$.}
\label{tab:musique-consistency}
\end{table}

The modest gains over baselines stem from (i) the lightweight, dated nature of SimCSE-BERT (circa 2021), which constrains fine-grained semantic capture, and (ii) the binary ``useful''/``noise'' labels inadequately capturing nuanced real-world document interrelations. 

\subsection{Ablation Studies on \texorpdfstring{$\tau$}{tau}}

We conduct ablation studies on the Musique dataset with GPT-4o-mini-2024-07-18 (top-$k$ = 20 and 100, noise = 40\%), evaluating the initial cluster count ($\tau$) across three independent trials. We report the mean and unbiased standard deviation of F1 scores and API call counts, with $\Lambda$ fixed for consistency. The results, presented in Table~\ref{tab:ablation-musique}, demonstrate stable performance across a wide range of $\tau$, affirming the robustness of our design.

\section{Conclusion}  
In this study, we design an efficient dynamic clustering algorithm and apply compression techniques to exploit fine-grained relationships between documents. Our method \textbf{EDC\textsuperscript{2}-RAG} enhances evidence quality by filtering noise and capturing detailed document relationships, achieving consistent performance improvements on three Hallucination-Detection datasets and three KQA datasets, thus demonstrating the strong robustness and broad applicability of our method. Extensive evaluations show that our approach outperforms competitive baselines across multiple metrics and model backbones.

\section*{Limitations}  

Our study has several limitations:  
1) Due to time constraints, we did not validate the generalization ability of our method on more datasets and base models.  
2) Using compression technique incurs some API consumption, but these costs are within an acceptable range. See Appendix~\ref{sec:api_costs} for details.

\section*{Acknowledgements}
This work is supported by  the National Natural Science Foundation of China (62372260, 62276152), and Wuxi Research Institute of Applied Technologies, Tsinghua University. Weizhi Ma is also supported by Beijing Nova Program.
\bibliography{anthology,custom}

\appendix

\section*{Appendix}

\section{API costs and Latency Control}
\label{sec:api_costs}

\paragraph{API Cost Evaluation.} 
To better understand the overhead introduced by different RAG compression strategies, we evaluate API token consumption using the tiktoken.encoding\_for\_model("gpt-3.5-turbo") tokenizer, which closely approximates OpenAI's official billing. Costs are computed based on the pricing of \texttt{gpt-4o-mini-2024-07-18}: 
\$0.15 per million input tokens and \$0.60 per million output tokens. 
We report results on the Musique dataset with $k=10$ and $k=100$ under the noise-free setting, and compare our method against RALM, Long Agent, and Chunk Compression. 
The key metric is the total API usage cost (input + output) across the full pipeline, including both document processing and final answering.

\begin{table}[h]
\centering
\small
\setlength{\tabcolsep}{2.5pt}
\begin{tabular}{lcccc}
\toprule
 & RALM & Chunk C. & Long Agent & \textbf{Ours} \\
\midrule
\multicolumn{5}{c}{\textcolor{gray}{\textbf{$k=10$, noise=0}}} \\
Avg Input   & 1388.45  & 2233.03  & 1843.42  & 2155.10 \\
Avg Output  & 34.97    & 740.70   & 223.73   & 553.29 \\
API Cost  & 2.29 & 7.79 & 4.11 & 6.55 \\
Rel. Cost   & 1.00     & 3.40     & 1.79     & 2.86 \\
\midrule
\multicolumn{5}{c}{\textcolor{gray}{\textbf{$k=100$, noise=0}}} \\
Avg Input   & 13542.94 & 20317.25 & 14406.18 & 14926.17 \\
Avg Output  & 38.89    & 6026.16  & 395.58   & 1212.89 \\
API Cost & 20.55 & 66.63 & 23.98 & 30.12 \\
Rel. Cost   & 1.00     & 3.24     & 1.17     & 1.46 \\
\bottomrule
\end{tabular}
\caption{API cost ($\times10^{-4}$) comparison on Musique under different $k$ settings.}
\end{table}

\paragraph{Cost Analysis.}
Our method achieves strong cost control, especially in large $k$ settings, for two main reasons: 
(1) one-time document access ensures bounded input-token cost, and 
(2) query-aware cluster-based compression balances relevance and brevity, avoiding the excessive output tokens incurred by Chunk Compression. 
In low-$k$ or noise-free settings, our cost is slightly higher than RALM and Long Agent. However, in such scenarios the total token usage is inherently small and noise is minimal (thus outside the target scenario of our method), making the overhead acceptable.

\paragraph{Efficiency Analysis.}
Our method is also efficient in runtime. 
We employ \texttt{SimCSE-BERT} (110M) as a lightweight encoder, and each document is encoded only once. 
The clustering step adds negligible overhead, and all summarization steps are \textbf{fully parallelizable}. 
In practice, this leads to wall-clock latency even lower than a single RALM query. 
These characteristics are consistent with our design goal of being \textbf{efficient}, as emphasized in the paper title.

\section{Implementation Details}
\label{sec:implementation details}


\subsection{Knowledge-QA Datasets and Retrieval Setup}
\label{sec:knowledge-qa-datasets-intro}
Knowledge Question Answering (KQA) datasets are essential resources for evaluating a model's ability to perform knowledge reasoning and question-answering tasks. These datasets typically rely on external knowledge bases (e.g., knowledge graphs or text corpora) and design questions to test the model's ability to retrieve information from the knowledge base and perform reasoning. In this work, we used three widely adopted datasets~\cite{yu-etal-2024-chain,lv2024coarse}: WebQ~\cite{berant2013semantic} (single-hop), and 2WikiMultiHopQA~\cite{ho-etal-2020-constructing} (hereafter referred to as 2Wiki) plus Musique~\cite{trivedi-etal-2022-musique} (both multi-hop).

WebQ is constructed by collecting questions posed by users in Google Suggest, with answers primarily based on the Freebase knowledge graph. The dataset is designed to test the model's ability to retrieve answers from structured knowledge bases while understanding natural language questions.  

2WikiMultiHopQA is a multi-hop question answering dataset automatically constructed from Wikipedia. Each question requires reasoning over two or more Wikipedia articles to arrive at the correct answer. It is designed to test a model's ability to perform compositional reasoning and handle longer context chains compared to single-hop datasets.

Musique is a multi-hop QA dataset with complex, natural questions decomposed into multiple factoid subquestions. It is built from real queries and aligned with Wikipedia paragraphs, making it suitable for evaluating models on realistic multi-hop reasoning tasks that require integrating information across multiple documents.

In this setting, we follow prior work on retrieval-augmented generation (RAG)~\cite{lv2024coarse,yu-etal-2024-chain,gao-etal-2023-enabling}, using the DPR retriever~\cite{karpukhin-etal-2020-dense} with the 2018 Wikipedia snapshot as the retrieval corpus, where each document contains approximately 100 words. For the three KQA datasets---WebQ, 2Wiki, and MuSiQue---we retrieve the top 1000 relevant documents for each test question. We apply string matching to identify whether each document contains the gold answer. A question is included in our final test set only if it has at least 100 documents with the answer (\textit{has\_answer}) and 100 without. This filtering yields test sets of approximately 400, 400, and 100 queries for WebQ, 2Wiki, and MuSiQue, respectively.

To build noisy retrieval scenarios, we inject the retrieved irrelevant documents into the retrieved set at controlled noise ratios. Document order is determined by similarity to the query. We vary the number of retrieved documents (top-$k$) from 5 to 100 and evaluate performance across different noise levels (0\% to 100\%) using the F1 score as the metric. The clustering threshold $\tau$ is set to 3 to balance document compression quality and API cost.

To evaluate the capability of our method in handling redundancy, we selected the $k$ documents when each question was associated with top-$20$ documents. The remaining $20-k$ documents were rewritten using ChatGPT. We define the redundancy rate as  

\[
r = \frac{20-k}{20}
\]

and construct datasets with redundancy rates of $r = 0.2,\ 0.4,\ 0.6,\ 0.8,\ \text{and}\ 0.95$ , corresponding to $k = 16,\ 12,\ 8,\ 4,\ \text{and}\ 1$ respectively.

\subsection{Hallucination Detection Datasets and Retrieval Setup}  
\label{sec:Hallucination-Detection-datasets-intro}
Fact-checking (Hallucination Detection) is a natural language processing task aimed at verifying the truthfulness and accuracy of generated or stated content. Specifically, it involves determining whether a given piece of generated text (often machine-generated, such as summaries, answers, translations, etc.) or statement is truthful, partially truthful, or false based on available information sources (i.e., containing "hallucinations" or erroneous content). We conducted experiments on three widely used fact-checking tasks: the FELM World Knowledge Subset~\cite{chen2023felm}, the WikiBio GPT-3 Dataset~\cite{manakul-etal-2023-selfcheckgpt}, and the HaluEval Dataset~\cite{li-etal-2023-halueval}.  

These datasets were constructed leveraging the generative capabilities of large language models. Researchers design a series of tasks or scenarios, collected model-generated content, and annotate it using domain-specific background knowledge. Specifically, the datasets include various examples of model outputs, which are manually labeled to classify their truthfulness. Labels indicate whether the content is truthful, partially truthful, or entirely false (in this work, partially truthful and false are treated as false). This method not only captures potential issues in model-generated content but also provides high-quality benchmark datasets for evaluating models' fact-checking capabilities. Below is a sample question.  

\begin{table}[h]  
\centering  
\scalebox{0.85}{  
\begin{tabular}{p{1.1\linewidth}}  
\hline  
\#Knowledge\#: The nine-mile byway starts south of Morehead, Kentucky and can be accessed by U.S. Highway 60. Morehead is a home rule-class city located along US 60 (the historic Midland Trail) and Interstate 64 in Rowan County, Kentucky, in the United States. \\  
\#Question\#: What U.S Highway gives access to Zilpo Road, and is also known as Midland Trail? \\  
\hdashline  
\#Right Answer\#: U.S. Highway 60 \\  
\#Hallucinated Answer\#: U.S. Highway 70 \\  
\hline  
\end{tabular}  
}  
\caption{A sample question from the HaluEval Dataset.}  
\label{tab:9}  
\end{table}  

For the FELM World Knowledge Subset and WikiBio GPT-3 Dataset, the queries are statements. The retrieval corpus consisted of an October 2023 snapshot of Wikipedia from CEG~\cite{li-etal-2024-citation}, and the retriever used is SimCSE Bert~\cite{gao-etal-2021-simcse}. The evaluation metric is Balanced Accuracy (Balanced-Acc).  

For the HaluEval Dataset, the retrieval corpus and setup were similar to those in other works~\cite{karpukhin-etal-2020-dense,gao-etal-2023-enabling}, employing a 2018 snapshot of Wikipedia and a state-of-the-art BERT-based retriever, All-mpnet-base-v2\footnote{\url{https://huggingface.co/sentence-transformers/all-mpnet-base-v2}}. The evaluation metric is Accuracy (Acc).  

In this scenario, due to the lack of a unified retrieval paradigm or specifically constructed retrieval corpus for such datasets, the contribution of documents to answering questions was inherently limited. We cap the number of retrieved documents at 10. Since the number of documents is small, \(\tau\) is set to 1 here to help the LLM summarize the documents more effectively.





\subsection{Detailed Introduction of Baselines} 
\label{sec:baselines}
The baselines for FELM include: 1) prompts enhanced with Chain-of-Thought (CoT) reasoning~\cite{kojima2022large}, 2) prompts augmented with hyperlinks to reference documents, and 3) prompts supplemented by human-annotated reference documents~\cite{chen2023felm}.

The baselines for WikiBio GPT-3 comprise: 1) HalluDetector\cite{wang-etal-2023-hallucination}, which leverages external knowledge sources along with a dedicated classification model and a Naive Bayes classifier to identify hallucinations, and 2) Focus\cite{zhang-etal-2023-enhancing-uncertainty}, which employs a multi-stage decision-making framework combining both pre-retrieval and task-specific classifiers.
\section{Prompts Used in Our Experiments}

\subsection{Hallucination Detection Datasets}

\subsubsection{FELM \& HaluEval}

\begin{tcolorbox}[colback=gray!10, colframe=black!50, title={Prompt of Compression}, breakable]

\textbf{\#\#Instruction\#\#}:

You are an AI assistant specializing in information extraction. Your task is to analyze a given statement and a set of related documents, and extract only the directly relevant information.\\[0.5em]
\textbf{\#\#Extraction Guidelines\#\#}:

- Identify key points, evidence, or details that **directly support, refute, or elaborate** on the statement.

- Ensure that the extracted content is **concise, objective, verifiable, and directly traceable** to the original documents.

- **Do not make inferences or draw conclusions** beyond what is explicitly stated.

- If the documents contain **no relevant information**, respond with **No content to extract.**\\[0.5em]
\textbf{\#\#Example Output Format\#\#}:

\texttt{\{few-shots\}}\\[0.5em]
\textbf{\#\#Statement\#\#}:

\texttt{\{query\}}\\[0.5em]
\textbf{\#\#Documents\#\#}:

\texttt{\{docs\}}\\[0.5em]
\textbf{\#\#Extracted Information\#\#}:
\end{tcolorbox}

\begin{tcolorbox}[colback=gray!10, colframe=black!50, title={Eval Prompt of HaluEval}, breakable]

\textbf{\#\#Instruction\#\#}:

I want you to act as an answer judge. Given a question, two answers, and related knowledge, your objective is to select the best and correct answer without hallucination and non-factual information.

You should try your best to select the best and correct answer. If the two answers are the same, you can choose one randomly. If both answers are incorrect, choose the better one. You MUST select an answer from the two provided answers.

Think step by step. Give your reasoning first and then output your choice. Output in the following format:

"\#Reasoning\#: Your Reasoning

\#Choice\#: "X"".

"X" should only be either "Answer 1" or "Answer 2", rather than specific answer content. \\[0.5em]
\textbf{\#\#Knowledge\#\#}:

\texttt{\{knowledge\}}\\[0.5em]
\textbf{\#\#Question\#\#}:

\texttt{\{question\}}\\[0.5em]
\textbf{\#\#Answer 1\#\#}:

\texttt{\{answer 1\}}\\[0.5em]
\textbf{\#\#Answer 2\#\#}:

\texttt{\{answer 2\}}

\end{tcolorbox}

\subsubsection{WikiBio GPT-3}

\begin{tcolorbox}[colback=gray!10, colframe=black!50, title={Prompt of Compression}, breakable]
\textbf{\#\#Instruction\#\#}:

You have been provided with a statement about \texttt{\{a person\}} and a collection of related documents. Your task is to extract relevant information from these documents that directly supports, refutes, or elaborates on the given statement. 

Focus on identifying key points, evidence, or details that are clearly connected to the statement. Ensure the extracted content is concise, directly relevant, and maintains the context of the original documents.

The extracted content must be objective, verifiable, and directly traceable to the original documents. Avoid making inferences or drawing conclusions based on the extracted content.

If you find that the documents contain no relevant information, please output "No content to extract". Below is an example. \\[0.5em]
\texttt{\{One shot\}}\\[0.5em]
\textbf{\#\#Person\#\#}:

\texttt{\{person\}}\\[0.5em]
\textbf{\#\#Statement\#\#}:

\texttt{\{query\}}\\[0.5em]
\textbf{\#\#Documents\#\#}:

\texttt{\{docs\}}\\[0.5em]
\textbf{\#\#Extracted Information\#\#}:

\end{tcolorbox}

\begin{tcolorbox}[colback=gray!10, colframe=black!50, title={Prompt of Evaluation}, breakable]

\textbf{\#\#Instruction\#\#}:

Assess whether the given statement about \texttt{\{a person\}} contains factual errors or not with the help of the reference docs.

If you believe given statement contains factual errors, your answer should be "Nonfactual"; if there is no factual error in this statement, your answer should be "Factual". This means that the answer is "Nonfactual" only if there are some factual errors in the given statement. When there is no factual judgment in the given statement or the given statement has no clear meaning, your answer should be "Factual". At the same time, please consider all aspects of the given statement thoroughly during the evaluation and avoid focusing excessively on any single factual aspect. Any factual errors should be considered.

Reference docs can be classified into three types: documents that support the response segment as "Nonfactual", documents that support the response segment as "Factual", and documents that provide supplementary or explanatory information for the response segment. Please consider these documents comprehensively when answering. 

Think it step by step. Give your "Reasoning" first and then output the "Answer". \\[0.5em]
\textbf{\#\#Statement\#\#}:

\texttt{\{statement\}} \\[0.5em]
\textbf{\#\#Reference docs\#\#}:

\texttt{\{passage\}} \\[0.5em]
\textbf{\#\#Output\#\#}:

\end{tcolorbox}

\subsection{Knowledge-QA Datasets}

















The prompts used for compression and generation in KQA tasks are shown below. These prompts differ from those used in previous datasets because we aim to elicit more informative chunks by having the model respond to the question first. This approach encourages the model to provide supporting evidence, which we then use to extract and compress relevant information. In contrast, directly prompting the model to summarize often leads it to provide answers directly without grounding them in the source content. If there is no strong formatting requirement, the quality of the LLM’s responses remains stable; however, if strict formatting requirements are imposed, the response quality drops sharply, causing a significant decline in performance. Accordingly, during the final generation stage, we also have the model consider these outputted answers and their corresponding evidence. The model integrates all the evidence to select the most appropriate answer.

\begin{tcolorbox}[colback=gray!10, colframe=black!50, title={Prompt of Summarization}, breakable]
\label{prompt:kqa_compression}
\textbf{\#\#Instruction\#\#}:

Please refer to the following text and answer the following question, \textbf{providing supporting evidence}. \\[0.5em]
\textbf{\#\#Question\#\#}:

\texttt{\{question\}} \\[0.5em]
\textbf{\#\#Reference text\#\#}:

\texttt{\{docs\}} \\[0.5em]
\textbf{\#\#Answer\#\#}:

\end{tcolorbox}

\begin{tcolorbox}[colback=gray!10, colframe=black!50, title={Prompt of Response}, breakable]
\label{prompt:kqa_answer}
\textbf{\#\#Task\#\#}:

Analyze the following set of candidate answers to a question and select the single most consistent/plausible answer based on majority consensus and logical coherence. \\[0.5em]
\textbf{\#\#Instructions\#\#}:

1. Carefully compare all candidate answers.

2. Identify the core factual claims or entities in each answer.

3. Group semantically equivalent answers (e.g., "1990", "the year 1990", "nineteen ninety").

4. Select the answer that:
    - Appears most frequently in the candidate set
    - Has strong internal consistency (no self-contradictions)
    
5. If multiple answers have equal validity, prefer the most specific and concise one. \\[0.5em]
\textbf{\#\#Format Requirements\#\#}:

Reasoning: Concise justification for selection.

Selected\_Answer:... \\[0.5em]

Below is an example.

Candidate Answers: 
["Paris", "The capital is Paris", "France", "paris", "It's Paris in France"]

Question: What is the capital of France?

Expected Response:

Reasoning: 4/5 answers directly state 'Paris'. While 'France' is incorrect alone, the most frequent and unambiguous consensus is 'Paris'
Selected\_Answer: Paris \\[0.5em]
\textbf{\#\#Candidate Answers\#\#}:

\texttt{\{answers\}} \\[0.5em]
\textbf{\#\#Question\#\#}:

\texttt{\{question\}} \\[0.5em]

\end{tcolorbox}






\section{Additional Experimental Results} 
\subsection{Experiments on Open-Source Models}

Additional experiments are conducted using Qwen-3-8B in think mode on the TwoWiki dataset under a noise rate of 0\%, constrained by available computational resources. These experiments, summarized in Table~\ref{tab:twowiki-supplement}, utilized only this 8B model. The results reveal a notable performance gap compared to closed-source LLMs, attributable to the limited summarization and evidence-filtering capabilities of smaller models.

\begin{table}[h]
\centering
\renewcommand{\arraystretch}{1.1}
\setlength{\tabcolsep}{6pt}
\begin{tabular}{l c c}
    \toprule
    Top-$k$ & RALM & Ours (Qwen-3-8B) \\
    \midrule
    5   & 66.96 & 60.33 \\
    10  & 72.39 & 67.71 \\
    20  & 73.90 & 75.64 \\
    30  & 78.44 & 71.01 \\
    50  & 80.76 & 69.88 \\
    70  & 80.30 & 72.17 \\
    100 & 81.56 & 71.18 \\
    \bottomrule
\end{tabular}
\caption{Performance comparison on TwoWiki dataset (noise rate 0\%) using Qwen-3-8B in think mode.}
\label{tab:twowiki-supplement}
\end{table}

We anticipate improved outcomes with larger open-source models and intend to incorporate corresponding experiments in future iterations, subject to resource availability.

\subsection{Additional Experimental Results on Noise Resistence} 

Tables~\ref{tab:noise_topk_20} summarizes performance under varying noise levels with Top-$k=20$.

\begin{table*}[ht]
\centering
\small
\setlength{\tabcolsep}{6pt}
\renewcommand{\arraystretch}{1.2}
\begin{tabular}{p{2.0cm} p{3.5cm} c c c c c c c}
\toprule
\textbf{Dataset} & \textbf{Method} 
& \multicolumn{7}{c}{\textbf{Noise Rates (\%) at Top-$k$=20}} \\
\cmidrule(lr){3-9}
& & 0 & 20 & 40 & 60 & 80 & 100 & Avg \\
\midrule
\multicolumn{9}{c}{\textcolor{gray}{\textbf{gpt-3.5-turbo-1106}}} \\
\multirow{4}{*}{MusiQue} 
& Vanilla RALM      & 74.75 & 77.82 & 78.07 & 74.92 & 74.42 & 74.30 & 75.71 \\
& Chunk Compression & 74.15 & 75.38 & 77.70 & \underline{78.01} & 71.89 & \underline{76.08} & 75.54 \\
& Long Agent        & \textbf{84.21} & \underline{83.41} & \textbf{79.02} & 76.12 & \underline{78.91} & 75.78 & \underline{79.58} \\
& \textbf{Ours}              & \underline{82.55} & \textbf{85.50} & \underline{78.28} & \textbf{83.58} & \textbf{82.53} & \textbf{79.88} & \textbf{82.05} \\[2pt]
\hdashline
\multirow{4}{*}{WebQ}
& Vanilla RALM      & 90.07 & 89.62 & 90.12 & 90.14 & \textbf{90.06} & 86.36 & 89.40 \\
& Chunk Compression & \underline{90.77} & 89.68 & 90.03 & \underline{90.79} & \underline{89.68} & 87.64 & 89.77 \\
& Long Agent        & 90.49 & \textbf{91.91} & \underline{90.54} & 89.46 & 88.81 & \textbf{\textbf{87.91}} & \underline{89.85} \\
& \textbf{Ours}              & \textbf{90.79} & \underline{91.87} & \textbf{90.75} & \textbf{91.00} & 89.23 & \underline{87.87} & \textbf{90.25} \\[2pt]
\hdashline
\multirow{4}{*}{2Wiki}
& Vanilla RALM      & \textbf{77.51} & 71.48 & 71.84 & \underline{68.40} & 67.57 & 66.01 & 70.47 \\
& Chunk Compression & 72.41 & 71.52 & 71.06 & 68.13 & \underline{69.75} & \underline{67.2}8 & 70.03 \\
& Long Agent        & 76.06 & \textbf{77.05} & \underline{74.20} & 71.07 & 69.35 & 66.99 & \underline{72.45} \\
& \textbf{Ours}              & \underline{76.20} & \underline{76.66} & \textbf{76.75} & \textbf{72.43} & \textbf{72.92} & \textbf{68.99}& \textbf{73.99} \\
\midrule
\multicolumn{9}{c}{\textcolor{gray}{\textbf{gpt-4o-mini-2024-07-18}}} \\
\multirow{4}{*}{MusiQue}
& Vanilla RALM      & 77.78 & 73.39 & 76.25 & 68.08 & 65.42 & 70.32 & 71.87 \\
& Chunk Compression & 75.67 & 75.33 & \underline{76.82} & 75.29 & 67.41 & 68.26 & 73.13 \\
& RAPTOR & 72.07 & \underline{78.46} & 75.95 & 71.15 & \underline{76.64} & 70.78 & 74.18 \\
& Long Agent        & \underline{80.43} & 76.67 & 72.50 & \underline{77.69} & 73.93 & \textbf{78.05} & \underline{76.55} \\
& \textbf{Ours}              & \textbf{81.71} & \textbf{80.44} & \textbf{81.10} & \textbf{78.98} & \textbf{77.50} & \underline{74.91} & \textbf{79.11} \\[2pt]
\hdashline
\multirow{4}{*}{WebQ}
& Vanilla RALM      & 85.07 & 89.89 & \underline{90.82} & 88.70 & 88.27 & 85.20 & 87.99 \\
& Chunk Compression & 90.77 & \underline{90.49} & 90.08 & \textbf{90.53} & \textbf{89.40} & \textbf{86.98} & \underline{89.71} \\
& Long Agent        & \textbf{91.94} & \textbf{91.49} & \textbf{90.86} & \underline{90.13} & \underline{88.60} & 86.79 & \textbf{89.80} \\
& \textbf{Ours} & \underline{91.89} & 90.36 & 90.76 & 89.43 & 88.40 & \underline{86.90} & 89.62 \\[2pt]
\hdashline
\multirow{4}{*}{2Wiki}
& Vanilla RALM      & \textbf{73.84} & \underline{73.03} & \underline{71.43} & \underline{69.03} & \textbf{67.53} & \textbf{60.88} & \textbf{69.29} \\
& Chunk Compression & 69.24 & 68.63 & 67.84 & 68.45 & 66.12 & 59.14 & 66.51 \\
& Long Agent        & 71.33 & \textbf{73.32} & 70.52 & 64.27 & 62.69 & 57.29 & 66.57 \\
& \textbf{Ours}              & \underline{72.86} & 71.92 & \textbf{72.58} & \textbf{69.60} & \underline{66.44} & \textbf{60.88} & \underline{69.05} \\
\bottomrule
\end{tabular}
\caption{Comparison of F1 scores under different noise levels at Top-$k$=20 on MusiQue, WebQ, and 2Wiki datasets for multiple retrieval methods.}
\label{tab:noise_topk_20}
\end{table*}
\end{document}